\documentclass[letterpaper]{article} 
\usepackage{aaai25}  
\usepackage{times}  
\usepackage{helvet}  
\usepackage{courier}  
\usepackage[hyphens]{url}  
\usepackage{graphicx} 
\usepackage{natbib}  
\usepackage{caption} 
\usepackage{algorithm}
\usepackage{algorithmic}
\usepackage{tabularx}
\newcolumntype{C}{>{\centering\arraybackslash}X}

\newcolumntype{Z}{>{\centering\let\newline\\\arraybackslash\hspace{0pt}}X}
\usepackage{newfloat}
\usepackage{listings}
\DeclareCaptionStyle{ruled}{labelfont=normalfont,labelsep=colon,strut=off} 
\floatstyle{ruled}
\newfloat{listing}{tb}{lst}{}
\floatname{listing}{Listing}
\title{GTG: Generalizable Trajectory Generation Model for Urban Mobility}

\author {
    Jingyuan Wang \textsuperscript{\rm 1, \rm 2, \rm 3, \thanks{Corresponding author}},
    Yujing Lin\textsuperscript{\rm 1},
    Yudong Li\textsuperscript{\rm 1}
}
\affiliations {
    \textsuperscript{\rm 1}School of Computer Science and Engineering, Beihang University, Beijing, China\\
    \textsuperscript{\rm 2}MIIT Key Laboratory of Data Intelligence and Management, Beihang University, Beijing, China\\
    \textsuperscript{\rm 3}School of Economics and Management, Beihang University, Beijing, China
}

\usepackage{bibentry}

\usepackage{amsmath}
\usepackage{bm}
\usepackage{subfig}
\usepackage{booktabs}
\usepackage{multirow}
\usepackage{threeparttable}
\usepackage{makecell}

\newcommand{\name}{GTG~}
\newtheorem{definition}{Definition}

\begin{document}
\maketitle
\vspace{-30pt}
\begin{abstract}
Trajectory data mining is crucial for smart city management. However, collecting large-scale trajectory datasets is challenging due to factors such as commercial conflicts and privacy regulations. Therefore, we urgently need trajectory generation techniques to address this issue. 
Existing trajectory generation methods rely on the global road network structure of cities. When the road network structure changes, these methods are often not transferable to other cities. In fact, there exist invariant mobility patterns between different cities: 1) People prefer paths with the minimal travel cost; 2) The travel cost of roads has an invariant relationship with the topological features of the road network. Based on the above insight, this paper proposes a \textbf{G}eneralizable \textbf{T}rajectory \textbf{G}eneration model (GTG). The model consists of three parts: 1) Extracting city-invariant road representation based on \textit{Space Syntax} method; 2) Cross-city travel cost prediction through disentangled adversarial training; 3) Travel preference learning by shortest path search and preference update. By learning invariant movement patterns, the model is capable of generating trajectories in new cities. Experiments on three  datasets demonstrates that our model significantly outperforms existing models in terms of generalization ability.

\end{abstract}

\begin{links}
\link{Code}{https://github.com/lyd1881310/GTG}
\end{links}

\section{Introduction}
\label{sec:introduction}
Trajectory data is essential for intelligent transportation (ITS) ~\cite{wu2019learning, liu2024full}, smart cities ~\cite{hettige2024airphynet, ji2022precision}, and location-based services (LBS) ~\cite{lbs2018, lbs2021dgeye}. Trajectory generation is key to addressing the issue of insufficient trajectory data, especially in scenarios involving privacy protection and conflicts of commercial interests. 

Urban trajectory generation is an active research field, and different types of methods have been proposed. These methods can be divided into two categories: knowledge-driven methods and data-driven methods~\cite{kong2023mobility}.

\textbf{Knowledge-driven methods} generate trajectories based on empirical and statistical patterns of human mobility, including the Gravity Model~\cite{gravity_1946}, Intervening Opportunities Model~\cite{interventing}, and EPR model~\cite{epr_1, epr_2}. These methods typically analyze human behavior using coarse-grained grids and derive physical priors with practical significance. By empirically summarizing human mobility behavior, these methods have achieved notable results in macroscopic traffic trajectory simulations. However, to capture human mobility behavior with finer granularity, data-driven approaches are increasingly emphasized.

\textbf{Data-driven methods} utilize deep neural networks to capture complex mobility patterns in trajectory data. 
Based on different architectures, these algorithms can be categorized into Seq2Seq-based~\cite{Seq2Seq}, GAN-based~\cite{SeqGAN}, VAE-based~\cite{volunteer, SVAE}, and Diffusion-based~\cite{difftraj} methods.  
With the advancements in data collection and management~\cite{manage_chen2015efficient, manage_ding2018ultraman}, data-driven methods have been widely applied. Although they have achieved significant success by learning from large datasets, they also lead to data dependency. 

Knowledge-driven methods require less trajectory data but struggle to capture complex mobility patterns. In contrast, data-driven methods, while offering better performance, face challenges in generalizing to new cities. In data-driven models, changes in a city’s topological structure can prevent the effective transfer of previously learned road representations to the new network. 

\begin{figure}[t]
    \centering
    \includegraphics[width=0.39\textwidth] 
    {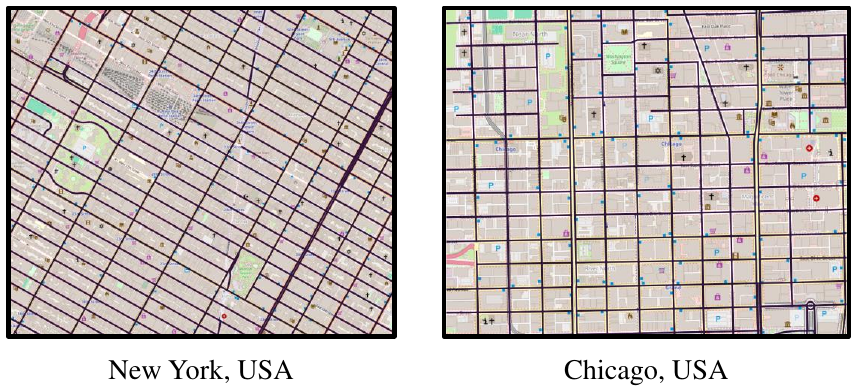}
    \caption{
        Similar local topological structures in New York (left) and Chicago (right). 
    }
    \label{fig:local_topology}
\end{figure}

The key to achieving cross-city trajectory generation is capturing the invariant human mobility patterns across different urban environments. Based on existing research ~\cite{pref2019, pref2021}, we have the following insights about the invariant human mobility patterns.
(\romannumeral1) People in different cities have similar travel preferences. Generally, people prefer paths with the minimal travel costs. By learning the combination of travel costs, we can generate trajectories that align with human preferences.
(\romannumeral2) The topological structure of the road network influences the functionality, usage frequency, and congestion levels of roads, thus determining the travel costs. Although the global road network structures differ between cities, there exists similar local topological structures. (see Figure~\ref{fig:local_topology}), making transfer learning possible.

Based on the above insights, we propose a cross-city trajectory generation model that combines shortest path search and deep learning. 
This model use \textit{Space Syntax} methods to extract topological features, employ a disentangled adversarial domain adaptation algorithm to learn invariant topological representations and predict travel costs, and finally learn human travel preferences to generate trajectories.

Our main contributions can be summarized in the following three points.
\begin{itemize}
    \item We combine \textit{Space Syntax} method with the advantages of deep neural networks to capture the invariant human mobility patterns across cities.
    \item We propose a novel cost prediction and preference learning method based on invariant topological features, and integrate it with the shortest path search algorithm to achieve cross-city trajectory generation. Our approach addresses the issue of cross-city generalization ability in trajectory generation.
    \item Experiments in multiple scenarios, datasets, and evaluation metrics show that our method has a generalization capability that far exceeds that of the baseline model.
\end{itemize}

\section{Preliminaries}
\label{sec:preliminaries}

\subsection{Definitions}
\begin{definition}[Road Network] Road network is represented as a graph $\mathcal{G = \langle R,E\rangle}$, where $\mathcal{R} = \{r_i\mid i = 1, 2,\cdots,N\}$ is the road segment set and $\mathcal E=\{e_{ij}\}$ is the edge set. $e_{ij} = 1$ if $r_i$ is adjacent to $r_j$, otherwise $e_{ij} = 0$. 
\end{definition}

\begin{definition}[Trajectory] A trajectory is a series of consecutive road segments $\tau = (r_1,r_2,\cdots,r_{l})$, where $l$ denotes the length of the trajectory. A dataset $\mathcal{T} = \{\tau_i \mid \tau_1, \tau_2, \cdots, \tau_C\}$ is formed by $C$ trajectories. \end{definition}

\subsection{Problem Statement}

Given a source city with a road network $\mathcal{G}^{(src)}$ and a trajectory dataset $\mathcal{T}^{(src)}$, the objective is to train a model $\mathcal{F}$ with parameters $\theta$ to generate a new trajectory dataset $\hat{\mathcal{T}}^{(tgt)}$ for a target city with a road network $\mathcal{G}^{(tgt)}$ but no trajectory data. The model is trained using the source city data as follows
\begin{equation}
\theta^{(src)} = \arg \min_{\theta} \mathcal{L} \left( \mathcal{F} \left( \mathcal{G}^{(src)}; \theta \right), \mathcal{T}^{(src)} \right),
\end{equation}
where $\mathcal{L}$ represents the optimization objective. The well-trained model is then applied to the target city to generate a new trajectory dataset
\begin{equation}
\hat{\mathcal{T}}^{(tgt)} = \mathcal{F} \left(
\mathcal{G}^{(tgt)}; \theta^{(src)}
\right).
\end{equation}
The primary objective is to ensure that the generated dataset $\hat{\mathcal{T}}^{(tgt)}$ is similar to the real dataset $\mathcal{T}^{(tgt)}$.
\section{Methodology}
\label{sec:methodology}

\begin{figure}[tbp]
    \centering
    \includegraphics[width=0.46\textwidth] 
    {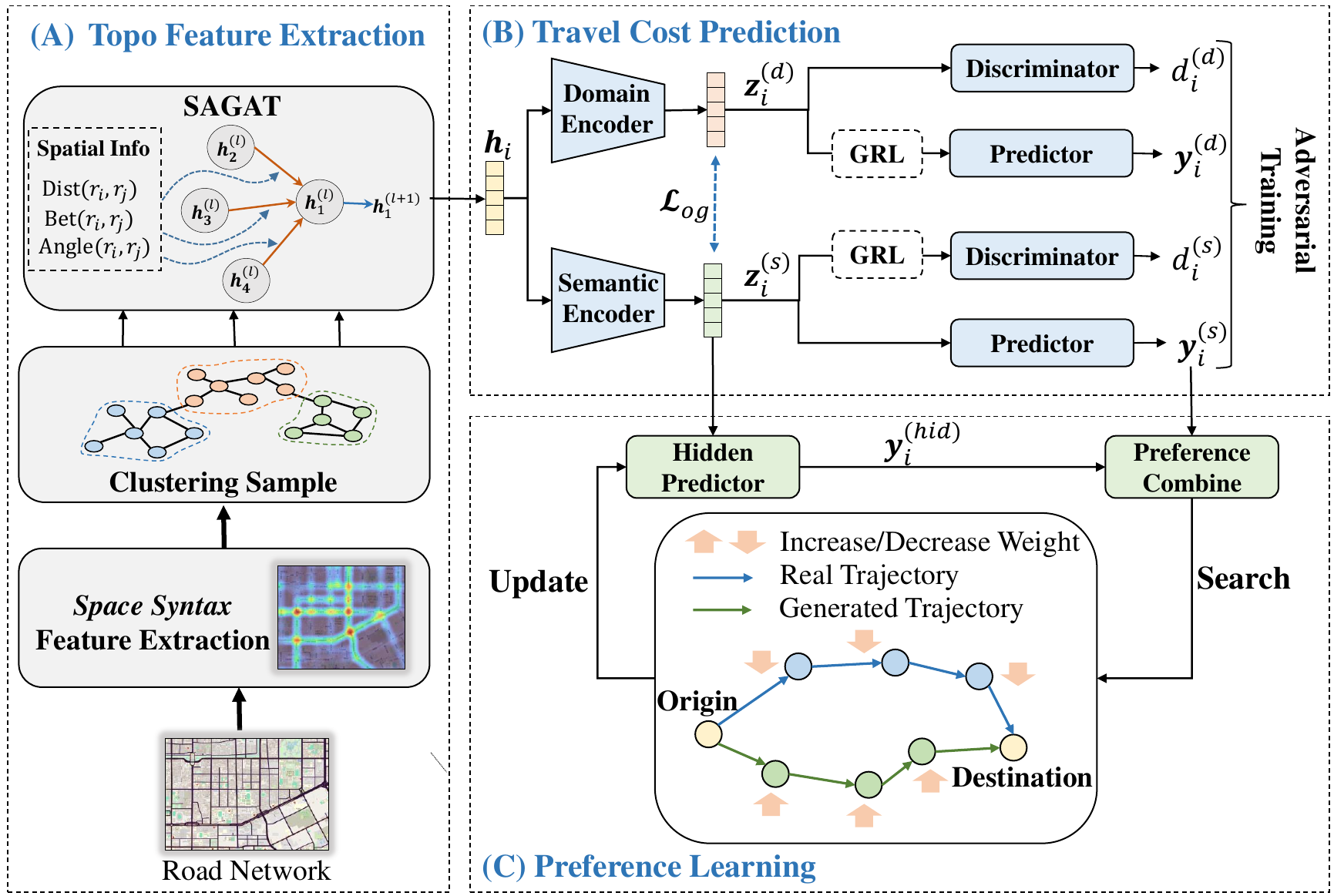}
    \caption{
        Overview of the framework
    }
    \label{fig:framework}
\end{figure}

\subsection{Framework}

As is shown in Figure~\ref{fig:framework}, our trajectory generation framework consists of three modules: Topological Feature Extraction, Travel Cost Prediction and Preference Learning. 
The training is performed using trajectory data in the source city and road network of both source city and target city. Trajectories are generated by inferring the preference values of each road segment and searching the shortest path.

\subsection{Topological Feature Extraction}
\subsubsection{\textit{Space Syntax} Feature Extraction}
\textit{Space Syntax}~\cite{hillier1976space} is a theory for analyzing and understanding spatial structure of cities and buildings. 
We use four types of concepts in \textit{Space Syntax} to describe the topological features of the road network.

\textbf{Total Depth} refers to the sum of the step depth (SD) from a given road segment to all other road segments within a certain range. SD is the minimum number of steps required to reach other target segment from a starting road segment.
\begin{equation}
    x_i^{(td)} = \sum_{j\neq i} \text{Len}(\text{ShortestPath}(r_i, r_j)).
\end{equation}
    
\textbf{Integration} measures the centrality of a road segment within the entire road network. The integration for road segment $r_i$ can be formulated as 
\begin{equation}
    x_i^{(in)} = \frac{\text{NC}(r_i)^2}{x_i^{(td)}}, 
\end{equation}
where $\text{NC}(r_i)$ refers to the total number of road segments that need to be traversed to reach all other road segments starting from $r_i$.

\textbf{Connectivity} indicates the number of directly connected neighbors to a road segment, which can be calculated as
\begin{equation}
    x_i^{(co)} = \text{Degree}(r_i) = \sum_j a_{ij}.
\end{equation}
If $r_i$ is adjacent to $r_j$, $a_{ij}$ equals $1$; otherwise, it equals $0$.

\textbf{Choice}, also known as Betweenness, reflects how often a road segment is likely to be encountered when moving through the space. The Choice for road segment $r_i$ is calculated by counting how many times it lies on the shortest paths between all pairs of segments $r_j$ and $r_k$, as follows
\begin{equation}
    x_i^{(ch)} = \sum_{j,k}{\delta_{ijk}}, 
\end{equation}
where $\delta_{ijk}$ indicates whether the shortest path from $r_j$ to $r_k$ passes $r_i$, formulated as 
\begin{equation}
    \delta_{ijk} = 
    \begin{cases}
        1, & \text{if} \  r_i \in \text{ShortestPath}(r_j, r_k), \\
        0, & \text{other}.
    \end{cases}
\end{equation}

Finally, The \textit{Space Syntax} features and the basic features of road segments, including length $x_i^{(le)}$, type $x_i^{(tp)}$ and direction $x_i^{(dr)}$, are concatenated as
\begin{equation}
    \bm{x}_{i} = x_i^{(td)} \| x_i^{(in)} \| x_i^{(co)} \| x_i^{(ch)} \| x_i^{(le)} \| x_i^{(tp)} \| x_i^{(dr)} \| t,
\end{equation}
where time slices index $t$ is incorporated to account for the varying travel costs of roads over time. Discrete variables are encoded through the embedding layer.

\subsubsection{Topological Feature Aggregation}
The raw road segment features are aggregated by GNNs to obtain representations with richer topological information.

Inductive GCN ~\cite{inductive_gcn} methods address generalization performance issues through subgraph learning. Sampling a variety of sub-graph samples can help further improve performance, but it is not possible to sample the entire graph due to computational efficiency. A compromise is to use the Metis algorithm, which partitions the entire road network into $K$ subgraphs, and $k$ of them are randomly sampled to form a training batch~\cite{ClusterGCN}, as follows
\begin{equation}
\begin{gathered}
    \left\{ \mathcal{\widetilde{G}}_1,  \mathcal{\widetilde{G}}_2, \cdots, \mathcal{\widetilde{G}}_K \right\} = \text{METIS} \left( \mathcal{G}, K \right), \\
    \mathcal{G}_k = \text{RandomSample} \left( \left\{ \mathcal{\widetilde{G}}_1,  \mathcal{\widetilde{G}}_2, \cdots, \mathcal{\widetilde{G}}_K \right\}, k \right).
\end{gathered}
\end{equation}

Next, Spatial Aware Graph Attention Networks (SAGAT) is designed to process the subgraph samples and obtain the aggregated representations. 
The SAGAT is a GATv2 ~\cite{GATv2} network in which we integrate spatial relationships between road segments to enhance the model's spatial awareness. This process could be formulated as 
\begin{equation}
    \left\{ \bm{h}_i \Big| r_i \in \mathcal{G}_k \right\} = \text{SAGAT} \left( \mathcal{G}_k \right).
\end{equation}

SAGAT is composed of multiple GAT layers stacked together, where the input of the first layer is a linear transformation of the features.
\begin{equation}
    \bm{h}_{i}^{(0)} = \text{MLP}(\bm{x}_{i}).
\end{equation} 
For the layer $l + 1$, the attention weights are calculated with the output of last layer $\bm{h}_{i}^{(l)}$, as follows
\begin{equation}
\begin{gathered}
    \alpha_{ij}^{(l+1)} = \frac{\exp{ \left( e_{ij}^{(l)} \right)}}{\sum_{j'\in \mathcal{N}_i}\exp \left( e_{ij'}^{(l)} \right)}, \\
    e_{ij}^{(l)} 
    = \bm{a}^\top \sigma \left( \bm{W}_s \bm{h}_i^{(l)} + \bm{W}_t \bm{h}_j^{(l)} + \bm{W}_e  \bm{s}_{ij} \right),
\end{gathered}
\end{equation}
where $\bm{a}$, $\bm{W}_s$, $\bm{W}_t$, $\bm{W}_e$ are learnable parameters, $\mathcal{N}_i$ is the neighbors of the road segment $r_i$ and $\sigma$ is LeaklyReLU function.
$\bm{s}_{ij}$ represents the spatial relationships between road segment $r_i$ and $r_j$, formulated as
\begin{equation}
    \bm{s}_{ij} =  \text{Bet}(r_i, r_j)\| \text{Angle}(r_i, r_j)\| \text{Dist}(r_i, r_j),
\end{equation}
where $\text{Bet}(r_i, r_j)$ is the Betweenness of road segment pair $r_i$ and $r_j$, which is defined as the ratio of the number of shortest paths that traverse $r_i$ and $r_j$ to the total number of shortest paths in the entire network. $\text{Angle}(r_i, r_j)$ represents the turning angle, and $\text{Dist}(r_i, r_j)$ represents the travel distance from the center of road segment.

The output of layer $l + 1$ is the weighted aggregation of the representations of neighboring nodes, as follows
\begin{equation}
    \bm{h}^{(l+1)}_{i} = \sigma \left(\sum_{j \in \mathcal{N}_i} \alpha_{ij}^{(l+1)} \bm{h}^{(l)}_{j}\right) + \bm{h}_{i}^{(l)},
\end{equation}
where $\sigma$ is the ReLU function. 

\subsection{Travel Cost Prediction with Disentangled Representation}

After aggregating the topological features, we aim to predict the travel cost based on the representation of the road segments while ensuring good generalization to the target city.

The difference in representation distribution between source and target cities reducing the model’s generalization ability. To address this, disentangled learning and adversarial domain adaptation are used to create city-invariant representations. Building on ~\cite{disent_ijcai}, the method assumes road segment information is determined by two independent latent variables: a semantic latent variable $\bm{z}^{(s)}$ and a domain latent variable $\bm{z}^{(d)}$. These are extracted using a semantic and a domain encoder, respectively. $\bm{z}^{(s)}$ captures semantic information for predicting trajectory costs, while $\bm{z}^{(d)}$ contains city-specific domain information, as follows
\begin{equation}
    \label{equ:dis_encoder}
    \begin{split}
    \bm{z}_i^{(s)} &= \text{SemEncoder} \left( \bm{h}_i \right), \\
    \bm{z}_i^{(d)} &= \text{DomEncoder} \left( \bm{h}_i \right).
    \end{split}
\end{equation}

The adversarial domain adaptation technique is used to train these representations, incorporating a predictor for travel cost estimation and a discriminator for city identification.

\subsubsection{Travel Cost Prediction}

The travel cost, represented by the average travel time and speed of road segments during specific periods, is denoted as
\begin{equation}
    \bm{y}_i = \left\{ y_i^{(m)} \Big| m \in \{\text{time, speed}\} \right\},
\end{equation}
 where $m$ denotes the type of cost. Using the disentangled latent variable $\bm{z}_i$ as input (with superscripts omitted for simplicity), the travel cost prediction network is formulated as
\begin{equation}
\hat{\bm{y}}_i = \log \left( 1 + \exp \left( \text{MLP} \left(\bm{z}_i \right) \right) \right),
\end{equation}
The prediction loss function comprises a MSE loss and a Rank loss. The MSE loss is 
\begin{equation}
    \mathcal{L}_{mse} = \frac{1}{N_s} \sum_{i} \left \Vert \hat{\bm{y}}_i - \bm{y}_i \right \Vert ^2,
\end{equation}
where $N_s$ is the size of source city dataset.

\begin{figure}[t]
    \subfloat{\includegraphics[width=0.22\textwidth]{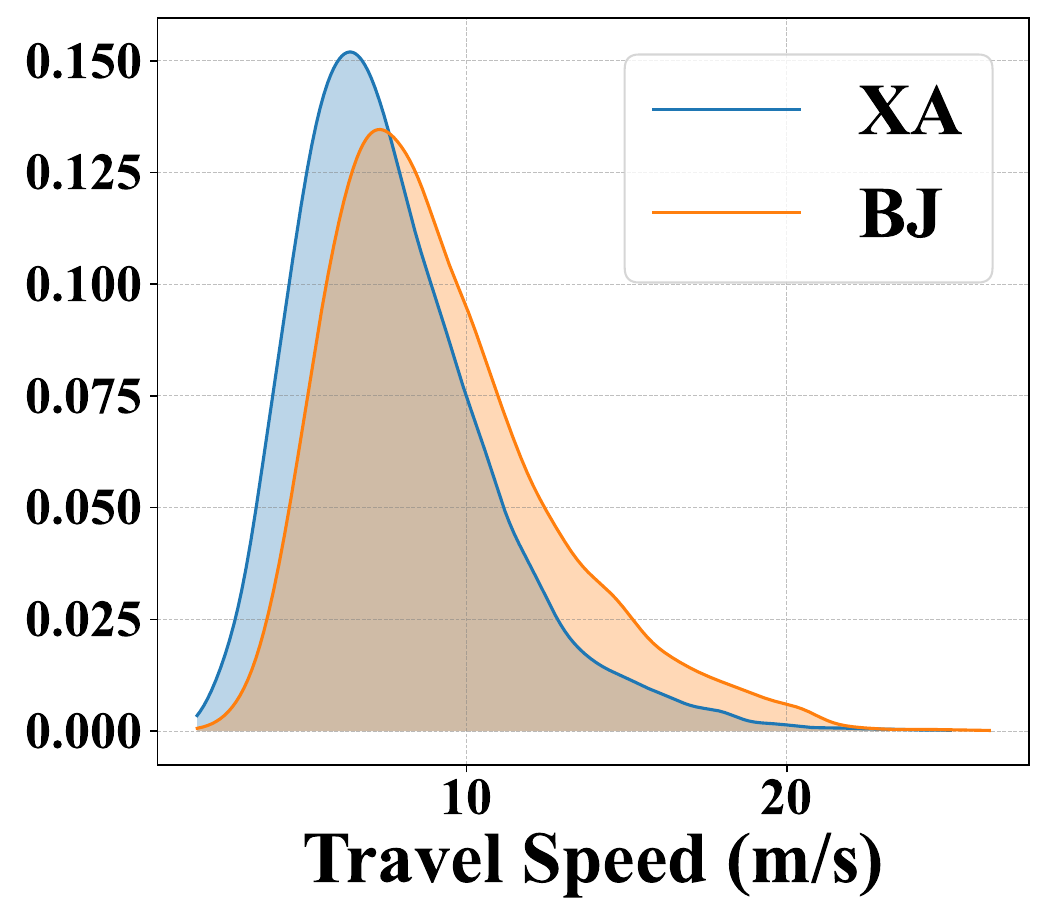}}
    \subfloat{\includegraphics[width=0.22\textwidth]{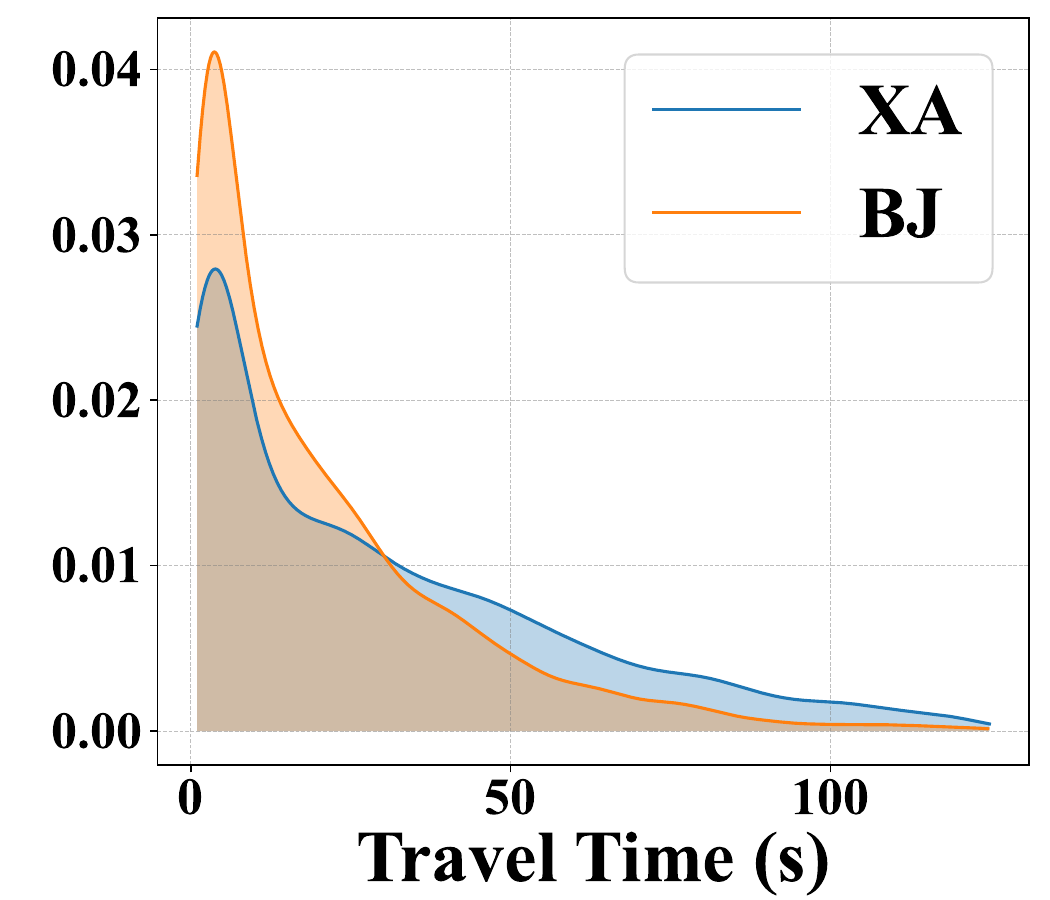}}
    \caption{Travel cost (speed and time) distribution in Xi'an and Beijing.}
    \label{fig:cost_distribution}
\end{figure}

Rank loss~\cite{ranknet} is motivated by the challenge of directly predicting absolute travel costs in the cross-city task, where potential biases across different cities can complicate predictions (see Figure \ref{fig:cost_distribution}). This approach emphasizes predicting relative rankings, which is generally less complex than estimating absolute values.

The probability that \(r_i\) has a higher travel cost than \(r_j\) is
\begin{equation}
\hat{q}_{ij}^{(m)} = \text{Sigmoid} \left( \hat{y}_i^{(m)} - \hat{y}_j^{(m)} \right).
\end{equation}
The actual ranking label of sample pair ($r_i$, $r_j$) is 
\begin{equation}
    q_{ij}^{(m)} = 
    \begin{cases}
        1, & \ y_i^{(m)} > y_j^{(m)}, \\
        0, & y_i^{(m)} < y_j^{(m)}.
    \end{cases}
\end{equation}
The binary cross-entropy loss is calculated as follows
\begin{equation}
\begin{gathered}
    l \left( \hat{q}, q \right) = - \left( q \log (\hat{q}) + (1 - q) \log (1 - \hat{q}) \right), \\
    \mathcal{L}_{rank} = \frac{1}{N_s^2} \sum_{i} \sum_{j} \sum_{m} l \left( \hat{q}_{ij}^{(m)}, q_{ij}^{(m)}\right).
\end{gathered}
\end{equation}
The overall loss for the travel cost prediction is
\begin{equation}
    \mathcal{L}_{pred} = \mathcal{L}_{mse} + \lambda_r \mathcal{L}_{rank},
\end{equation}
where $\lambda_r$ is a balance weight between the above two loss.

\subsubsection{Domain Discrimination}
To separate domain information from semantic information, we introduce a discriminator to predict the domain label of the road segment.
We extract subgraph sample from source city or target city and assign a domain label to each road segment, as follows
\begin{equation}
    d_{i} = 
    \begin{cases}
        1, & \ r_i \in \mathcal{G}^{(src)}, \\
        0, & \ r_i \in \mathcal{G}^{(tgt)}.
    \end{cases}
\end{equation}
Given the latent variable $\bm{z}_i$ as input, the domain discriminator is formulated as
\begin{equation}
    \hat{d}_i = \text{Sigmoid} \left( \text{MLP} \left( \bm{z}_i \right) \right).
\end{equation}
Binary cross entropy loss is used for this domain discrimination task, as follows
\begin{equation}
    \mathcal{L}_{dis} = - \frac{1}{N_s + N_t} \sum_i \left( d_i \log \hat{d}_i + ( 1 - d_i) \log ( 1 - \hat{d}_i) \right),
\end{equation}
where $N_s$ and $N_t$ represent the sizes of the source city and target city datasets, respectively.

\subsubsection{Disentangled Adversarial Training} Adversarial training is used to promote the information decoupling of $z_i^{(s)}$ and $z_i^{(s)}$. The ultimate objective for semantic latent variables is to minimize travel cost prediction loss while maximizing domain discrimination loss, whereas for domain latent variables, the goal is the opposite. By using different representations as input, the loss functions are calculated as follows
\begin{equation}
\begin{split}
\mathcal{L}_{total}^{(s)} = \mathcal{L}_{pred}\left(z_i^{(s)}\right) - \lambda_d \mathcal{L}_{dis}\left(z_i^{(s)}\right), \\
\mathcal{L}_{total}^{(d)} = \lambda_d \mathcal{L}_{dis}\left(z_i^{(d)}\right) - \mathcal{L}_{pred}\left(z_i^{(d)}\right).    
\end{split}
\end{equation}

As shown in Figure~\ref{fig:framework}, the GRL layer is used to negate the gradient to achieve efficient adversarial training, which performs an identity transformation in the forward propagation and negates the gradient in the back propagation.
In addition, the orthogonal loss is introduced to further reduce information coupling, calculated as follows
\begin{equation}
\mathcal{L}_{og} = \frac{1}{N_s + N_t}\sum_i \left( \frac{\bm{z}_i^{(s)} \cdot \bm{z}_i^{(d)}}{ \left\| \bm{z}_i^{(s)} \right\| \cdot \left\| \bm{z}_i^{(d)} \right\| } \right)^2.
\end{equation}
The total loss function for disentangled domain adaptation combines these losses
\begin{equation}
\mathcal{L}_{total} = \mathcal{L}_{total}^{(s)} + \mathcal{L}_{total}^{(d)} + \lambda_g \mathcal{L}_{og}.
\end{equation}
\subsection{Travel Preference Learning}

After completing the travel cost prediction, we can use the shortest path search algorithm to generate trajectories for the target city.

Shortest path search algorithms rely on fixed road cost factors, such as travel speed or time, for route planning. However, focusing on a single cost factor often fails to fully capture users’ actual travel preferences, which are influenced by more complex factors ~\cite{rideshare_chen2018price}. 

To address this, we propose modeling travel preferences as a combination of observable costs and hidden costs. Hidden costs, which account for harder-to-explain factors influencing human choices, are generated using a multi-layer perceptron (MLP) as follows
\begin{equation}
y_i^{(hid)} = \log \left( 1 + \exp \left( \text{MLP} (\bm{z}_i) \right) \right),
\end{equation}
where \(\bm{z}_i\) is a semantic latent variable.
We then estimate overall travel preference using a weighted combination of observable and hidden costs
\begin{equation}
\label{equ:combined_cost}
p(r_i) = \sum_{m} w^{(m)} y_i^{(m)} + y_i^{(hid)},
\end{equation}
where \(w^{(m)}\) are learnable weights. The smaller the value of \(p(r_i)\), the higher the preference for the road segment \(r_i\). We believe this method of combining preferences remains consistent across different cities.

The preference is learned by an unsupervised training. During training, we first randomly initialize the parameters and search for the shortest path. The shortest path found from $r_i$ to $r_j$ is denoted as $\hat{\tau}_{ij} = (r_i, \cdots, r_j)$, whose preference sum is 
\begin{equation}
    \hat{p} \left( \hat{\tau}_{ij} \right) = \sum_{r_k \in \hat{\tau}_{ij}} p(r_k).
\end{equation}
The sum of the preference values of the real trajectory ${\tau}_{ij}$ is
\begin{equation}
    p \left( \tau_{ij} \right) = \sum_{r_k \in \tau_{ij}} p(r_k).
\end{equation}
And then the loss function is formulated as 
\begin{equation}
    \mathcal{L}_{pref} = \frac{1}{\left| \mathcal{T}^{(src)} \right|} \sum_{\tau_{ij} \in \mathcal{T}^{(src)}} \left( p \left( \tau_{ij} \right) - \hat{p} \left( \hat{\tau}_{ij} \right)\right).
\end{equation}
Through iterative training, the model learns the invariant mapping relationship between travel preferences and various travel costs, which can then be applied to the target city to generate trajectory data. It is worth mentioning that assigning a cost to each road segment is similar to the MaxEnt IRL~\cite{max_ent}. Theoretical analysis can be found in the code repository.

\section{Experiments}
\begin{table*}[t]
\small
\centering
\fontsize{7}{8}\selectfont
\setlength{\tabcolsep}{8pt}
\begin{tabular}{c|c|cccccccccc|cc}
\toprule
Target & Metric & RW & DE & SE & TG & SG & SV & MS & TT & DT & VO & GTG$^{1}$ & GTG$^{2}$ \\
\midrule
\multirow{7}{*}{BJ} & Distance & 0.0434 & 0.1509 & 0.1662 & 0.2980 & 0.2320 & 0.0080 & 0.1826 & 0.0107 & 0.0173 & 0.1893 & \textbf{0.0006} & \underline{0.0006} \\ 
    & Radius & 0.1191 & 0.1332 & 0.1543 & 0.2180 & 0.2070 & 0.0098 & 0.1450 & 0.0071 & 0.0019 & 0.0518 & \textbf{0.0001} & \underline{0.0002} \\ 
    & LocFreq & 0.198 & 0.076 & 0.196 & 0.187 & 0.038 & 0.044 & 0.410 & 0.095 & 0.216 & 0.145 & \textbf{0.041} & \underline{0.043} \\ 
    & Hausdorff & 2.682 & 3.833 & 3.807 & 7.228 & 10.469 & 3.294 & 7.523 & 1.092 & 3.305 & 5.680 & \textbf{0.292} & \underline{0.315} \\ 
    & DTW & 51.54 & 58.88 & 58.49 & 185.17 & 149.49 & 61.99 & 145.38 & 24.27 & 74.26 & 89.82 & \textbf{4.89} & \underline{5.27} \\ 
    & EDT & 28.17 & 29.85 & 29.90 & 42.93 & 27.31 & 26.20 & 31.07 & 18.13 & 31.42 & 29.17 & \textbf{8.73} & \underline{9.17} \\ 
    & EDR & 0.813 & 0.904 & 0.906 & 0.820 & 0.903 & 0.751 & 0.942 & 0.423 & 0.889 & 0.850 & \textbf{0.190} & \underline{0.205} \\
\midrule
\multirow{7}{*}{XA} & Distance & 0.0524 & 0.2122 & 0.2233 & 0.4271 & 0.1369 & 0.0584 & 0.3018 & 0.0085 & 0.0386 & 0.1450 & \underline{0.0044} & \textbf{0.0040} \\ 
    & Radius & 0.0708 & 0.1422 & 0.1612 & 0.4199 & 0.0655 & 0.0268 & 0.1795 & 0.0011 & 0.0030 & 0.0588 & \textbf{0.0002} & \underline{0.0002} \\ 
    & LocFreq & 0.263 & 0.104 & 0.264 & 0.403 & 0.126 & 0.099 & 0.290 & 0.097 & 0.180 & 0.207 & \underline{0.042} & \textbf{0.040} \\ 
    & Hausdorff & 2.300 & 2.406 & 2.627 & 2.852 & 2.319 & 1.597 & 3.161 & 0.349 & 1.006 & 2.178 & \underline{0.188} & \textbf{0.187} \\ 
    & DTW & 34.02 & 36.35 & 38.04 & 56.34 & 26.20 & 17.66 & 30.69 & 5.28 & 12.09 & 23.52 & \textbf{2.32} & \underline{2.34} \\ 
    & EDT & 24.21 & 30.28 & 30.36 & 34.11 & 17.67 & 13.42 & 17.95 & 7.10 & 14.88 & 15.85 & \textbf{4.29} & \underline{4.23} \\ 
    & EDR & 0.818 & 0.888 & 0.893 & 0.711 & 0.808 & 0.595 & 0.871 & 0.243 & 0.668 & 0.680 & \underline{0.142} & \textbf{0.135} \\ 
\midrule
\multirow{7}{*}{CD} & Distance & 0.0646 & 0.2083 & 0.2069 & 0.3596 & 0.1420 & 0.0405 & 0.2560 & 0.0136 & 0.0440 & 0.1380 & \underline{0.0051} & \textbf{0.0049} \\ 
    & Radius & 0.0958 & 0.1127 & 0.0980 & 0.1259 & 0.0619 & 0.0221 & 0.1598 & 0.0017 & 0.0053 & 0.0162 & \textbf{0.0002} & \underline{0.0002} \\ 
    & LocFreq & 0.280 & 0.110 & 0.267 & 0.210 & 0.103 & 0.051 & 0.374 & 0.069 & 0.159 & 0.098 & \underline{0.027} & \textbf{0.026} \\ 
    & Hausdorff & 1.434 & 1.981 & 1.980 & 2.104 & 2.433 & 1.273 & 3.126 & 0.220 & 1.033 & 1.530 & \textbf{0.117} & \underline{0.125} \\ 
    & DTW & 18.87 & 24.82 & 24.32 & 28.75 & 28.65 & 13.94 & 35.44 & 2.54 & 13.52 & 15.62 & \textbf{1.19} & \underline{1.34} \\ 
    & EDT & 19.42 & 25.64 & 25.83 & 20.16 & 18.27 & 14.02 & 18.42 & 6.69 & 17.48 & 15.01 & \textbf{3.59} & \underline{3.63} \\ 
    & EDR & 0.778 & 0.865 & 0.869 & 0.720 & 0.801 & 0.620 & 0.900 & 0.213 & 0.679 & 0.657 & \textbf{0.109} & \underline{0.117} \\ 
\bottomrule
\end{tabular}
\caption{New City Trajectory Generation Results 
}
\label{tab:main_table}
\begin{tablenotes}
\footnotesize
\item [1] GTG$^1$ refers to the GTG model trained on XA for BJ, BJ for CD, and CD for XA. GTG$^2$ refers to the model trained on BJ for XA, XA for CD and CD for BJ.
\end{tablenotes}
\end{table*}

\subsection{Experimental Settings}
Experiments are conducted on three real trajectory datasets. The experiments consists of four parts: New City Trajectory Generation, Downstream Task Support, Target City Fine-tune and Ablation Study.

\subsubsection{Datasets and Preprocessing}
Three real-world trajectory datasets are used to evaluate the performance of our proposed method. These datasets were collected in the three cities, namely Beijing(BJ), Xi'an(XA) and Chengdu(CD).

Road network of the three cities are collected from the OpenStreetMap ~\cite{OpenStreetMap}, and road segment trajectories are obtained by performing map-matching algorithm ~\cite{fmm}.

\subsubsection{Comparative Baselines} 
Baseline models can be categorized into two types, knowledge-driven and data-driven.
\begin{itemize}
    \item Knowledge-Driven Methods:
    The knowledge-driven methods are manually proposed by researchers based on the analysis of trajectory data. Subsequently, they integrate proposed rules with a random walk algorithm to generate trajectory data. This type of baseline includes Random Walk~\cite{node2vec}(RW), Density-EPR~\cite{epr_1}(DE) and Spatial-EPR~\cite{epr_2}(SE).

    \item Data-Driven Methods: TrajGen~\cite{trajgen}(TG), SeqGAN~\cite{SeqGAN}(SG), SVAE~\cite{SVAE}(SV), MoveSim~\cite{MoveSim}(MS), TS-TrajGen~\cite{ts_trajgen}(TT), DiffTraj ~\cite{difftraj}(DT) and VOLUNTEER ~\cite{volunteer}(VO).
\end{itemize}

\subsubsection{Evaluation Metrics}
Evaluation metrics are divided into two categories, macroscopic and microscopic.
\begin{itemize}
    \item Macro metrics: we use the Jensen-Shannon divergence(JSD) to evaluate the similarity between the generated trajectory dataset and the real dataset across three statistical features: travel distance(Distance), radius of gyration(Radius), and road segment visit frequency(LocFreq). 
    \item Micro metrics: we calculate the sequence distance between the each generated trajectory and its corresponding real trajectory to evaluate similarity. We use four types of sequence distance: Hausdorff, DTW, EDT and EDR.
\end{itemize}

\subsection{New City Trajectory Generation Experiment}
\begin{figure}[t]
    \centering
    \subfloat[Real]{\includegraphics[width=0.15\textwidth]{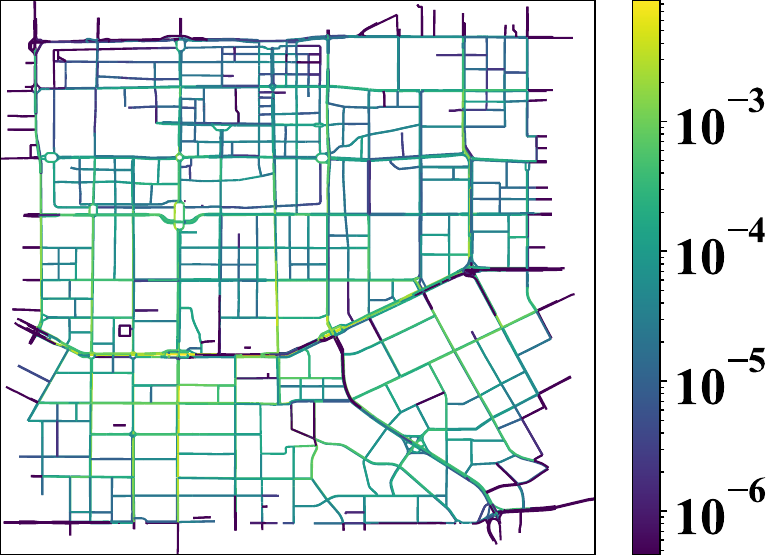}}
    \subfloat[GTG]{\includegraphics[width=0.15\textwidth]{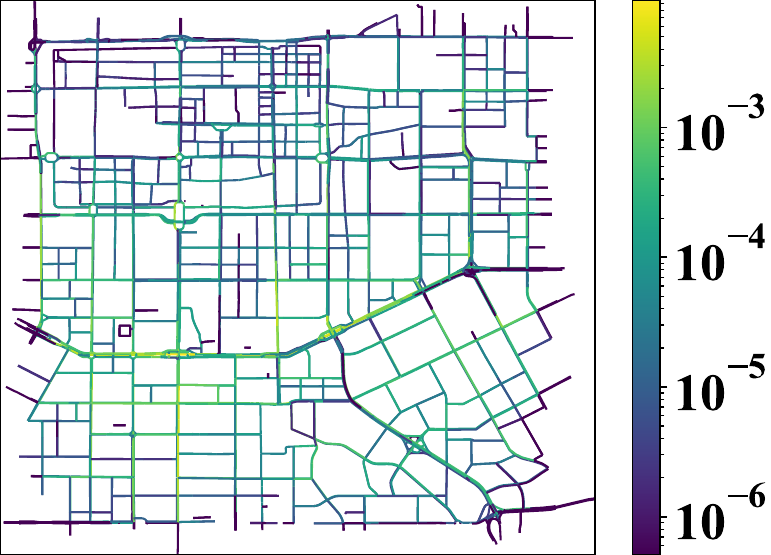}}  
    \subfloat[SV]{\includegraphics[width=0.15\textwidth]{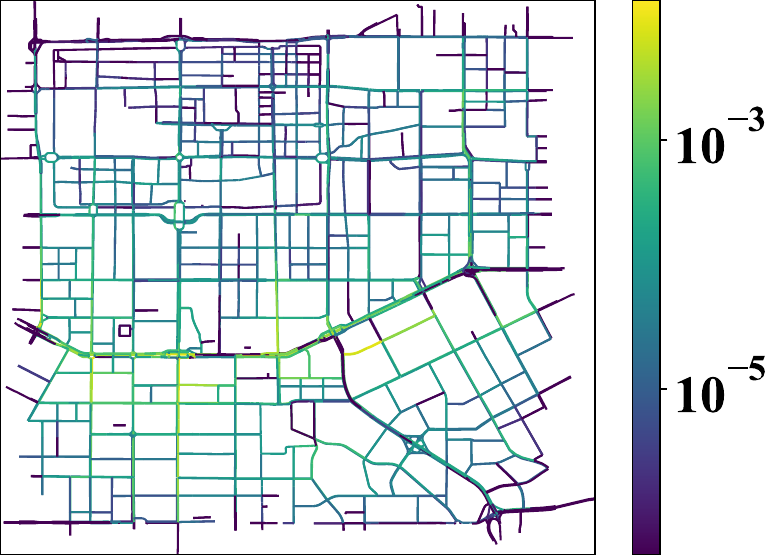}}

    \subfloat[DE]{\includegraphics[width=0.15\textwidth]{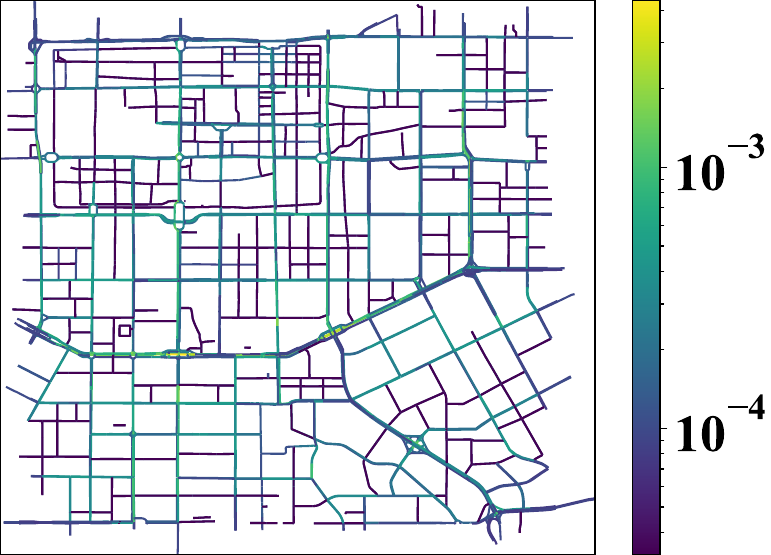}}
    \subfloat[DT]{\includegraphics[width=0.15\textwidth]{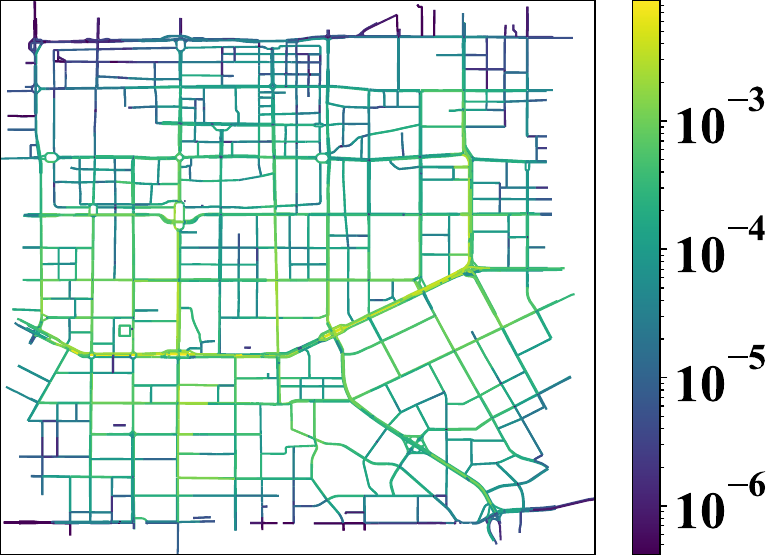}}
    \subfloat[VO]{\includegraphics[width=0.15\textwidth]{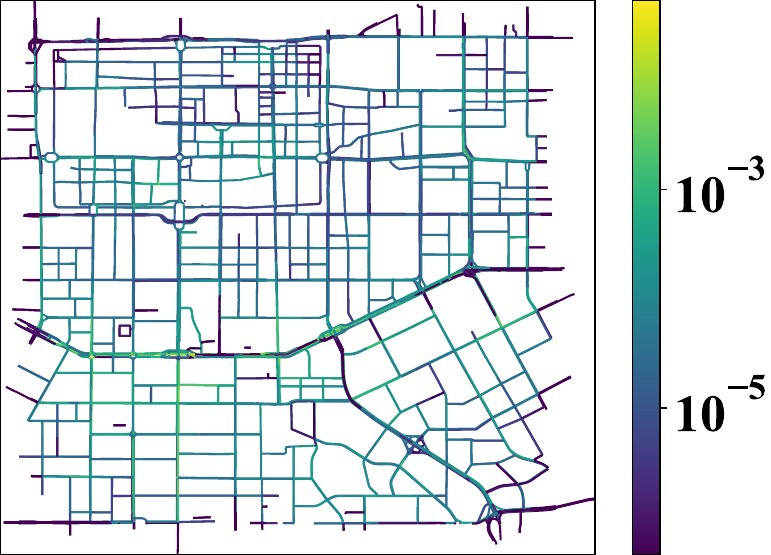}}

    \caption{The visualization of road segment visit frequency on the XA dataset, brighter color means higher frequency.}
    \label{fig:loc_freq_visualize}
\end{figure}
The overall performance result is shown in Table~\ref{tab:main_table}. In our experiments across each dataset, the best results are highlighted in bold, while the second-best results are underlined. 

The proposed method demonstrates superior performance compared to all baseline models on the three real-world trajectory datasets, evaluated from both macro and micro perspectives. Significant improvements are observed across all metrics with the proposed approach. Unlike other deep learning baseline methods that require training on the target city’s data to produce accurate trajectories, \name can operate effectively without training in target city. This demonstrates the model’s generalization capability.

Additionally, all evaluation metrics exhibit smaller values, suggesting that the generated trajectory data closely resembles the real trajectory data. Figure~\ref{fig:loc_freq_visualize} shows the generated datasets and real datasets for some baselines. \name achieves the highest similarity to the real dataset.

\subsection{Downstream Task Support}
\begin{table}[t]
  \centering
  \fontsize{7}{8}\selectfont
    \begin{tabularx}{\linewidth}{C|C|CCC|CCC|CCC}
      \toprule
      & \multicolumn{1}{c|}{\multirow{2}[4]{*}{Data}}  & \multicolumn{3}{c|}{BJ} & \multicolumn{3}{c|}{XA} & \multicolumn{3}{c}{CD} \\
       \cmidrule(lr){3-5} \cmidrule(lr){6-8} \cmidrule(lr){9-11}
      & & ACC & NDCG & MRR & ACC & NDCG & MRR & ACC & NDCG & MRR \\
      \midrule
    \multirow{12}{*}{\rotatebox{90}{DeepMove}} & Real & 0.81 & 0.88 & 0.86 & 0.88 & 0.94 & 0.93 & 0.89 & 0.95 & 0.93\\ 
    \cmidrule{2-11}
        & RW & 0.60 & 0.75 & 0.71 & 0.62 & 0.83 & 0.78 & 0.62 & 0.85 & 0.80\\ 
        & DE & 0.01 & 0.04 & 0.03 & 0.01 & 0.11 & 0.08 & 0.02 & 0.11 & 0.08\\ 
        & SE & 0.01 & 0.02 & 0.02 & 0.01 & 0.03 & 0.03 & 0.01 & 0.05 & 0.03\\ 
        & TG & 0.56 & 0.61 & 0.60 & 0.46 & 0.47 & 0.46 & 0.69 & 0.74 & 0.73\\ 
        & SG & 0.67 & 0.73 & 0.71 & 0.64 & 0.70 & 0.68 & 0.82 & 0.89 & 0.87\\ 
        & SV & 0.64 & 0.68 & 0.67 & 0.78 & 0.81 & 0.80 & 0.82 & 0.88 & 0.87\\ 
        & MS & 0.01 & 0.02 & 0.02 & 0.11 & 0.13 & 0.13 & 0.11 & 0.13 & 0.18\\ 
        & TT & 0.65 & 0.73 & 0.71 & 0.71 & 0.78 & 0.76 & 0.75 & 0.83 & 0.81\\ 
        & DT & 0.14 & 0.20 & 0.18 & 0.39 & 0.52 & 0.49 & 0.41 & 0.51 & 0.49\\ 
        & VO & 0.22 & 0.31 & 0.29 & 0.42 & 0.49 & 0.47 & 0.60 & 0.69 & 0.67\\ 
        \cmidrule{2-11}
        & \name$^1$ & \underline{0.72} & \underline{0.80} & \underline{0.78} & \textbf{0.82} & \underline{0.87} & \textbf{0.85} & \textbf{0.84} & \textbf{0.90} & \textbf{0.89}\\ 
        & \name$^2$ & \textbf{0.73} & \textbf{0.80} & \textbf{0.79} & \underline{0.80} & \textbf{0.85} & 0.84 & \underline{0.83} & \underline{0.89} & \underline{0.88}\\ 

      \midrule
      \multirow{12}{*}{\rotatebox{90}{LSTPM}} & Real & 0.85 & 0.93 & 0.91 & 0.90 & 0.96 & 0.95 & 0.89 & 0.96 & 0.94\\ 
        \cmidrule{2-11}
        & RW & 0.68 & 0.86 & 0.82 & 0.64 & 0.85 & 0.80 & 0.61 & 0.84 & 0.79\\ 
        & DE & 0.01 & 0.05 & 0.04 & 0.01 & 0.10 & 0.08 & 0.00 & 0.11 & 0.08\\ 
        & SE & 0.00 & 0.03 & 0.02 & 0.00 & 0.03 & 0.02 & 0.00 & 0.05 & 0.04\\ 
        & TG & 0.66 & 0.70 & 0.69 & 0.62 & 0.66 & 0.65 & 0.78 & 0.84 & 0.83\\ 
        & SG & \underline{0.80} & 0.87 & 0.85 & 0.75 & 0.82 & 0.81 & 0.85 & 0.93 & 0.91\\ 
        & SV & 0.79 & 0.86 & 0.84 & 0.83 & 0.88 & 0.87 & \underline{0.87} & \underline{0.93} & \underline{0.92}\\ 
        & MS & 0.04 & 0.05 & 0.05 & 0.13 & 0.16 & 0.15 & 0.15 & 0.17 & 0.17\\ 
        & TT & 0.74 & 0.85 & 0.83 & 0.77  & 0.85  & 0.84 & 0.80 & 0.89 & 0.87\\ 
        & DT & 0.17 & 0.26 & 0.23 & 0.43  & 0.56 & 0.53 & 0.45 & 0.58 & 0.55\\ 
        & VO & 0.30 & 0.38 & 0.36  & 0.49 & 0.56  & 0.54 & 0.64 & 0.73 & 0.71\\ 
        \cmidrule{2-11}
        & \name$^1$ & \textbf{0.80} & \textbf{0.88} & \textbf{0.87} & \textbf{0.86} & \textbf{0.91} & \textbf{0.90} & \textbf{0.87} & \textbf{0.94} & \textbf{0.92}\\ 
        & \name$^2$ & 0.78 & \underline{0.87} & \underline{0.60} & \underline{0.85} & \underline{0.90} & \underline{0.89} & 0.87 & 0.93 & 0.92\\ 
      \bottomrule
    \end{tabularx}%
  \caption{Downstream Task Support Experiment Result}
  \label{tab:downstream}%
\end{table}%

\begin{table}[t]
    \centering
    \fontsize{7}{8}\selectfont
    \setlength{\tabcolsep}{3pt}
    \begin{tabular}{c|c|ccc|cccc}
      \toprule
       City & \makecell{ \#Traj \\ ($10^3$)} & \makecell{ Distance \\ ($10^{-3}$)} & \makecell{ Radius \\ ($10^{-3}$)} &  LocFreq & Hausdorff &  DTW  &  EDT &  EDR \\
      \midrule
      \multirow{6}{*}{\rotatebox{90}{BJ $\to$ XA}} & 0.0 & 4.039 & 0.228 & 0.040 & 0.187 & 2.34 & 4.23 & 0.135 \\ 
        & 0.1 & 4.679 & 0.825 & 0.040 & 0.194 & 2.52 & 4.36 & 0.143 \\ 
        & 0.4 & 4.875 & 0.306 & 0.044 & 0.188 & 2.30 & 4.38 & 0.147 \\ 
        & 1.6 & 4.233 & 0.334 & 0.036 & 0.184 & 2.21 & 4.17 & 0.139 \\ 
        & 6.4 & 3.976 & 0.139 & 0.029 & 0.174 & 2.07 & 4.00 & 0.129 \\ 
        & 12.8 & 4.034 & 0.104 & 0.027 & 0.170 & 1.95 & 3.96 & 0.125 \\ 
      \midrule
      \multirow{6}{*}{\rotatebox{90}{CD $\to$ XA}} & 0.0 & 4.375 & 0.206 & 0.042 & 0.188 & 2.32 & 4.29 & 0.142 \\ 
        & 0.1 & 4.468 & 0.345 & 0.050 & 0.192 & 2.45 & 4.67 & 0.148 \\ 
        & 0.4 & 5.951 & 0.464 & 0.067 & 0.207 & 2.72 & 5.05 & 0.159 \\ 
        & 1.6 & 4.285 & 0.175 & 0.043 & 0.181 & 2.13 & 4.38 & 0.139 \\ 
        & 6.4 & 4.103 & 0.219 & 0.033 & 0.171 & 1.95 & 4.09 & 0.130 \\ 
        & 12.8 & 4.128 & 0.253 & 0.029 & 0.170 & 1.93 & 4.01 & 0.128 \\ 
      \bottomrule
    \end{tabular}%
    \caption{Target City Fine-tune Experiment Results in XA}
    \label{tab:finetune}%
\end{table}%
\begin{table}[t]
    \centering
    \fontsize{7}{8}\selectfont
    \setlength{\tabcolsep}{2.5pt}
    \begin{tabular}{c|c|ccc|cccc}
      \toprule
        City & \makecell{Method} & \makecell{Distance \\ ($10^{-3}$)} & \makecell{Radius \\ ($10^{-3}$)} & LocFreq & Hausdorff & DTW  & EDT & EDR \\
      \midrule
      \multirow{4}{*}{\rotatebox{90}{BJ $\to$ XA}} & w/o Cost & 19.171 & 5.066 & 0.119 & 0.264 & 3.78 & 8.51 & 0.260 \\ 
        & w/o Pref & 4.356 & 0.293 & 0.047 & 0.198 & 2.54 & 4.61 & 0.146 \\ 
        & w/o SS & 3.685 & 0.673 & 0.041 & 0.199 & 2.57 & 4.49 & 0.144 \\ 
        & \name & 4.039 & 0.228 & 0.040 & 0.187 & 2.34 & 4.23 & 0.135 \\ 
      \midrule
      \multirow{4}{*}{\rotatebox{90}{CD $\to$ XA}} & w/o Cost & 17.745 & 3.763 & 0.118 & 0.274 & 3.52 & 6.98 & 0.227 \\ 
        & w/o Pref & 4.641 & 1.052 & 0.062 & 0.219 & 3.00 & 5.03 & 0.164 \\ 
        & w/o SS & 5.261 & 2.817 & 0.059 & 0.228 & 3.15 & 4.95 & 0.163 \\ 
        & \name & 4.375 & 0.206 & 0.042 & 0.188 & 2.32 & 4.29 & 0.142 \\  
      \bottomrule
    \end{tabular}%
    \caption{Ablation Study Results in XA}
    \label{tab:ablation}%
\end{table}%
Trajectory data is frequently used in many downstream tasks, such as next-hop prediction ~\cite{DeepMove, LSTPM}, trajectory classification~\cite{traj_clf_wang2018cd, traj_chen2019real}, and trajectory recovery~\cite{traj_recover_wang2019deep}.
The downstream task experiment serves as a supplementary verification of the model’s generalization ability and underscores the practical significance of the trajectory generation task. In this experiment, the trajectory generation model is employed to produce pre-training data for the downstream task, thereby enhancing the performance.

Specifically, the trajectory next-location prediction task is selected as the downstream application. This task, which aims to predict the next location in a trajectory given several observed locations, is widely applicable in POI recommendation systems. The models DeepMove~\cite{DeepMove} and LSTPM~\cite{LSTPM}, implemented via LibCity~\cite{libcity}, are utilized for this purpose. These downstream models are trained using the trajectory data generated by the proposed model, and their performance is subsequently tested on real trajectory data.

The results, as presented in Table~\ref{tab:downstream}, are evaluated using three metrics: Accuracy (ACC), Normalized Discounted Cumulative Gain at 3 (NDCG@3), and Mean Reciprocal Rank at 3 (MRR@3). The performance of the data generated by \name in training downstream tasks is found to be second only to that of real data. The baseline model demonstrates inferior performance, indicating that the trajectory data generated by our approach is more effective in supporting downstream tasks.

\subsection{Target City Fine-tune}

Considering that the gradual collection of trajectory data is a more realistic application scenario, it would be helpful if the trajectory generation capability of the model could further adjust in the target city. We fine-tune the model using the trajectory of the target city to test its improved generation ability. Fine-tuning phase training include travel cost prediction and preference learning. The experimental results of fine-tuning in XA are shown in the Table~\ref{tab:finetune}.  The results in other cities can be found in our code repository.

From Table~\ref{tab:finetune}, we can see that using target city data for fine-tuning improves the model's performance. Before applying the model to a new city, collecting a small amount of trajectory data for fine-tuning can achieve good generation results without incurring excessive costs.

\subsection{Ablation Study}

To validate the effectiveness of submodules, we conduct the following ablation studies.

(a) w/o Cost: we removed the cost prediction module, which means that instead of using combined costs shown in formula ~\ref{equ:combined_cost}, we only use hidden costs as road weights to generate trajectories.

(b) w/o Pref: we removed the preference learning module, meaning that we now only use supervise learning to predict observable costs and use their sum as weights for trajectory generation. 

(c) w/o SS: we removed the \textit{Space Syntax} feature extraction module to test the impact of \textit{Space Syntax} features., which means that only the basic features of road segments are input into the model.

The ablation study results in target city XA are shown in Table~\ref{tab:ablation}. 
Upon the removal of the aforementioned submodules, a notable decline in the model's performance was observed, with the cost prediction module having the most pronounced impact. Ablation experiments that excluded the cost prediction and preference learning modules demonstrated that the integration of travel cost components more accurately captures the invariant travel patterns of humans. Additionally, experiments that removed the \textit{Space Syntax} feature extraction module revealed that \textit{Space Syntax} significantly contributes to the cross-city trajectory generation.

\section{Related Work}
Existing trajectory generation works can be divided into two categories: Knowledge-Driven methods and Data-Driven methods.

\textbf{Knowledge-driven methods} models human mobility based on prior knowledge and statistical data. Gravity Model~\cite{gravity_1946} and Intervening Opportunities Model~\cite{interventing} generate trajectories by modeling the relationship between inter-regional mobility intensity and population and economic data. The EPR model~\cite{epr_1, epr_2} regards trajectory generation as a process of multiple explorations and returns, and the probability of an individual visiting a location is related to regional attractiveness. This type of method emerged along with traditional modeling methods. It has a coarse granularity but has certain generalization capabilities.

\textbf{Data-driven methods} uses large amounts of trajectory data to train neural network models,  capturing complex travel patterns. According to different architectures, these algorithms can be divided into algorithms based on Seq2Seq ~\cite{Seq2Seq}, GAN ~\cite{SeqGAN}, VAE ~\cite{volunteer, SVAE}, and Diffusion ~\cite{difftraj}. The Seq2Seq method uses RNN, TCN, Transformer to model the movement state of individual, then predict the choice of individuals facing multiple optional road segments. As an improvement to Seq2Seq, Neural A* ~\cite{ts_trajgen} searches multiple trajectories simultaneously to provide the most likely trajectory.  Similarly, the VAE method models the individual's movement state and travel prefernece through latent variables. In order to better generate trajectories with composite human movement distribution, the GAN method leverage a discriminator to identify the authenticity of the trajectory. Different from the method of generating trajectory component by component, Diffusion innovatively treats the trajectory of human movement as a special texture and generates the entire trajectory at once using image generation. This type of method focuses on statistical learning and has generalization capabilities in the same city but cannot be generalized to other cities.

In other areas of urban data mining, such as traffic flow prediction  and OD (origin-destination) demand prediction ~\cite{predji2023spatio, predjiang2023pdformer, predwang2022traffic}, there have been numerous active studies on cross-city transfer learning. ~\cite{wang2018cross} calculates matching scores between regions using other data and forces the model to generate similar representations for matched region pairs by adding a matching regularization term. 
CrossTRes~\cite{jin2022selective} conducts transfer learning using data from multiple cities. To avoid negative transfer caused by data from dissimilar cities, this method introduces a weighted score for each city to weight the losses from multiple cities. Yu Zheng et al.~\cite{he2020human} proposed an OD transfer learning model primarily focused on generating travel demand for new cities. Research on tranfer learning in other fields inspired us to propose \name.

\section{Conclusion}

In this paper, we propose a novel, generalizable trajectory generation model that leverages invariant human mobility patterns. The model incorporates \textit{Space Syntax} theory as a feature input and innovatively applies methods such as inductive graph convolution and disentangled learning to capture these complex mobility patterns. Across a wide range of experimental scenarios, baseline models, and evaluation metrics, this method consistently and significantly outperforms the baselines, which demonstrates the model’s exceptional generalization ability. In the future work, we will further enhance the model by allowing trajectory generation without training on the target city’s road network.
\newpage

\section*{Acknowledgements}
Prof. Jingyuan Wang's work was supported by the National Natural Science Foundation of China (No. 72171013, 72222022, 72242101), and the Special Fund for Health Development Research of Beijing (2024-2G-30121). 

\bibliography{main}

\clearpage
\appendix 
\section{Supplementary Illustration}
\subsection{\textit{Space Syntax}}
\label{apdx:space_syntax}
\textit{Space Syntax} was proposed and developed by British architectural scholars Bill Hillier and Julienne Hanson at University College London (UCL) in the 1970s. Hillier et al. analyzed London's street network through spatial syntax, revealing the relationship between street structure and crime rate, and found that streets with high integration are more easily monitored and have lower crime rates. 

In general, \textit{Space Syntax} provides a unique perspective to understand and optimize spatial structure, reveals the profound impact of spatial configuration on human behavior and social interaction through quantitative analysis, and provides a scientific basis for urban and architectural design.

\begin{figure}[t]
    \centering
    \includegraphics[width=0.40\textwidth] 
    {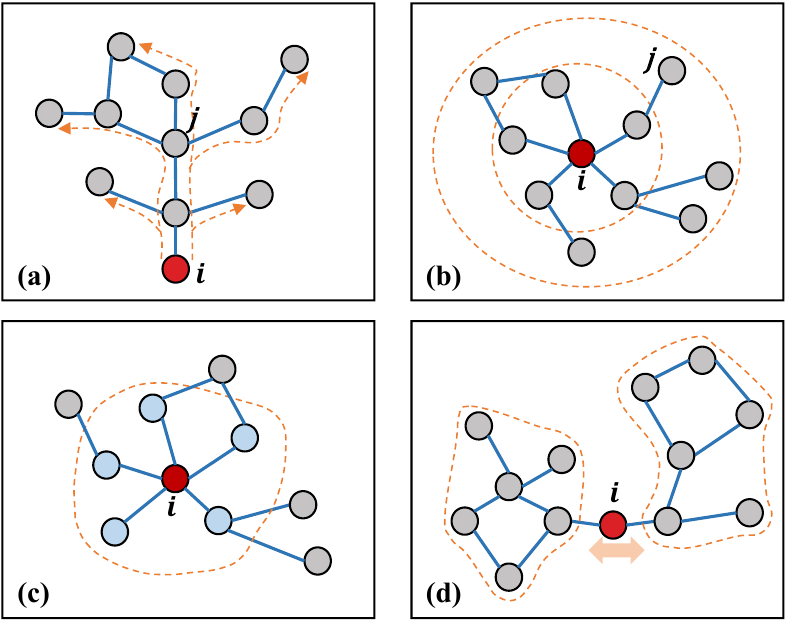}
    \caption{
        Four types of \textit{Space Syntax} concepts: (a) node $i$ has a larger Total Depth than node $j$; (b) node $i$ is in the center of network, with larger Integration than node $j$; (c) Connectivity is only related with neighborhood nodes, node $i$ 's Connectivity is $5$; (d) node $i$ is a transportation hub in this network, so it has larger Choice than other nodes.       
    }
    \label{fig:space_syntax}
\end{figure}

Four types of \textit{Space Syntax} features are applied in our model, and Figure ~\ref{fig:space_syntax} provides a vivid explanation of them.

\subsection{Max Entropy Inverse Reinforcement Learning}
\label{apdx:irl}
Below, we derive the preference learning process from the perspective of maximum entropy inverse reinforcement learning. 
We treat the negative value of road preference weights as rewards. The maximum entropy learning assumes that the reward function is set in such a way that the entropy of the probability of expert trajectories is maximized. Let $P(\tau|\omega)$ denote the probability of trajectory $\tau$ given the reward function parameters $\omega$. Then, maximum entropy inverse reinforcement learning solves the following problem under the feature matching constraint,
\begin{equation}
\begin{aligned}
    &\omega^{*} = \arg \max _\omega \sum_\tau -P(\tau|\omega)\log P(\tau | \omega) \\
    &\text{s.t.} \begin{cases}
    \sum_\tau P(\tau | \omega) \sum_i p(\tau_i) = \sum_{\tau\in \mathcal{T}} P(\tau | \omega) \sum_i p(\tau_i)  \\
    \sum_\tau P(\tau | \omega) = 1
    \end{cases}
\end{aligned}
\end{equation}
where $\mathcal{T}$ is the expert trajectory dataset, and $p(\tau_i)$ is the preference for $\tau_i$.

According to the assumptions in ~\cite{max_ent}, 
the above optimization problem is equivalent to solving the following maximum likelihood problem
\begin{equation}
\begin{gathered}
    \arg \max_\omega \sum_{\tau\in \mathcal{T}} \log P(\tau|\omega), \\ 
    P(\tau|\omega) =\frac{1}{Z(\omega)} \exp\left(-\sum_i p(\tau_i) \right), \\
    Z(\omega) =\sum_\tau \exp\left(-\sum_i p(\tau_i)\right). 
\end{gathered}
\end{equation}
The partition function $Z(\omega)$ represents the sum of the trajectory logit of the best strategies under the given $\omega$ setting.
The gradient of the likelihood function is calculated as follows
\begin{equation}
\begin{split}
    &\nabla_\omega \sum_{\tau\in \mathcal{T}} \log P(\tau|\omega) \\
    &=\nabla_\omega\left( -\sum_{\tau\in \mathcal{T}}\sum_i p(\tau_i)\right)  - |\mathcal{T}|\nabla_\omega\log Z(\omega),     
\end{split}
\end{equation}
Where the gradient of the second term is derived as follows
\begin{equation}
\begin{split}
    \nabla_\omega \log Z(\omega) &= \sum_\tau \frac{\exp \left( \sum_i -p(\tau_i) \right)}{Z(\omega)} \left(-\sum \nabla_\omega p(\tau_i, \omega) \right) \\ 
    &=-\sum_\tau P(\tau|\omega)\sum \nabla_\omega p(\tau_i)
\end{split}
\end{equation}
Since computing the entire space of possible trajectories is computationally infeasible, we use Monte Carlo sampling to approximate this gradient term. Specifically, in each iteration, we generate the shortest path trajectory $\hat{\tau}$, which corresponds to the real trajectory $\tau$. We then compute the rewards for both under the current preference settings and calculate the gradient.
\begin{equation}
    \nabla_\omega \log P(\tau|\omega) = \nabla_\omega \left( -\sum_i p(\tau_i) + \sum_{i} p(\hat \tau_i) \right)
\end{equation}
Therefore, maximizing the above likelihood function is equivalent to minimizing our loss function
\begin{equation}
    \mathcal{L}^{(pref)} = \frac{1}{\left| \mathcal{T}^{(src)} \right|} \sum_{\tau_{ij} \in \mathcal{T}^{(src)}} \left[ p \left( \tau_{ij} \right) - \hat{p} \left( \hat{\tau}_{ij} \right)\right].
\end{equation}

\section{Datasets}
\subsection{Datasets Description}
\label{apdx:data_desc}
The BJ dataset contains real GPS trajectory data of Beijing taxis from November 1 to 30, 2015 within Beijing’s Fourth Ring Roads, which is sampled every minute. The XA and CD datasets are originally released by DIDI Chuxing. The detailed information of the three datasets is shown in the Table~\ref{tab:dataset_statistic}. These datasets were chosen due to the number of trajectories and the time intensive collection.
\begin{table}[t]
  \centering
  \captionsetup{skip=5pt}
    \resizebox{\columnwidth}{!}{
        \begin{tabular}{c|ccc}
        \toprule
          Datastatistics & BJ & XA & CD \\
        \midrule
          Time Period & Nov. 2015 & Nov. 2018 & Nov. 2018 \\
          Trajectories & 2344762 & 804302 & 1252233 \\
          Road Segments & 14685 & 4147 & 3514 \\
          Average Hops & 24.0 & 13.4 & 14.4 \\
          Average Travel Distance(km) & 4.10 & 2.38 & 2.14 \\
        \bottomrule
        \end{tabular}
    }
    \caption{Datasets Statistics}
    \label{tab:dataset_statistic}
\end{table}

\subsection{Data Processing}
\label{apdx:data_proc}
To eliminate abnormal trajectories, we remove trajectories with lengths of less than 3 steps and trajectories with loops. We partitioned the trajectory datasets of each city into training and testing sets. The supervised training label, including travel time and travel speed for each road segment, are generated from the training set, while the testing set was used to evaluate the quality of the generated trajectories.

Our model needs to predict travel costs, therefore, we need to compute travel cost labels for each road segment based on trajectory dataset. The map-matching algorithm can output the travel time on each road segment for each trajectory. Because the travel cost of roads varies dynamically over time, we divide a day into 24 time slots and calculate the average travel time of roads within each hour. For each road segment, we collect all travel time samples within a time slot and filter out outliers exceeding $3 \sigma $. Finally, we calculate the average of the remaining samples as the label for the road segment's travel cost. After completing the data preprocessing, the basic information of the three datasets is shown in Table ~\ref{tab:dataset_statistic}. 

\section{Implementation Details}
\begin{table*}[t]
    \centering
    \resizebox{1.0\textwidth}{!}{
      \begin{tabular}{c|c|ccccccc|ccccccc}
      \toprule
        \multirow{2}{*}{Target} & \multirow{2}{*}{\makecell{\#Traj \\ ($\times 10^3$)}} & \multicolumn{7}{c|}{Source City XA for BJ, BJ for CD and CD for XA} & \multicolumn{7}{c}{Source City BJ for XA, XA for CD and CD for BJ} \\
        \cmidrule{3-16}
         &  & \makecell{Distance \\ ($\times 10^{-3}$)} & \makecell{Radius \\ ($\times 10^{-3}$) } & LocFreq & Hausdorff & DTW  & EDT & EDR & \makecell{Distance \\ ($\times 10^{-3}$)} & \makecell{Radius \\ ($\times 10^{-3}$)} & LocFreq & Hausdorff & DTW  & EDT & EDR \\
      \midrule

    \multirow{6}{*}{BJ} & 0.0 & 0.636 & 0.111 & 0.041 & 0.292 & 4.89 & 8.73 & 0.190 & 0.643 & 0.182 & 0.043 & 0.315 & 5.27 & 9.17 & 0.205 \\ 
        & 0.1 & 0.612 & 0.122 & 0.036 & 0.289 & 4.88 & 8.48 & 0.186 & 0.455 & 0.339 & 0.040 & 0.311 & 5.08 & 9.01 & 0.203 \\ 
        & 0.4 & 0.533 & 0.189 & 0.035 & 0.283 & 4.63 & 8.48 & 0.184 & 0.461 & 0.352 & 0.041 & 0.308 & 4.96 & 9.04 & 0.202 \\ 
        & 1.6 & 0.561 & 0.150 & 0.033 & 0.274 & 4.47 & 8.30 & 0.179 & 0.474 & 0.275 & 0.038 & 0.300 & 4.86 & 8.85 & 0.197 \\ 
        & 6.4 & 0.689 & 0.095 & 0.034 & 0.262 & 4.19 & 8.32 & 0.172 & 0.582 & 0.121 & 0.035 & 0.264 & 4.20 & 8.18 & 0.174 \\ 
        & 12.8 & 0.620 & 0.114 & 0.028 & 0.250 & 4.02 & 8.00 & 0.164 & 0.574 & 0.124 & 0.031 & 0.254 & 4.05 & 7.96 & 0.167 \\ 
    \midrule
    \multirow{6}{*}{XA} & 0.0 & 4.375 & 0.206 & 0.042 & 0.188 & 2.32 & 4.29 & 0.142 & 4.039 & 0.228 & 0.040 & 0.187 & 2.34 & 4.23 & 0.135 \\ 
        & 0.1 & 4.468 & 0.345 & 0.050 & 0.192 & 2.45 & 4.67 & 0.148 & 4.679 & 0.825 & 0.040 & 0.194 & 2.52 & 4.36 & 0.143 \\ 
        & 0.4 & 5.951 & 0.464 & 0.067 & 0.207 & 2.72 & 5.05 & 0.159 & 4.875 & 0.306 & 0.044 & 0.188 & 2.30 & 4.38 & 0.147 \\ 
        & 1.6 & 4.285 & 0.175 & 0.043 & 0.181 & 2.13 & 4.38 & 0.139 & 4.233 & 0.334 & 0.036 & 0.184 & 2.21 & 4.17 & 0.139 \\ 
        & 6.4 & 4.103 & 0.219 & 0.033 & 0.171 & 1.95 & 4.09 & 0.130 & 3.976 & 0.139 & 0.029 & 0.174 & 2.07 & 4.00 & 0.129 \\ 
        & 12.8 & 4.128 & 0.253 & 0.029 & 0.170 & 1.93 & 4.01 & 0.128 & 4.034 & 0.104 & 0.027 & 0.170 & 1.95 & 3.96 & 0.125 \\ 
    \midrule
    \multirow{6}{*}{CD} & 0.0 & 5.109 & 0.233 & 0.027 & 0.117 & 1.19 & 3.59 & 0.109 & 4.902 & 0.248 & 0.026 & 0.125 & 1.34 & 3.63 & 0.117 \\ 
        & 0.1 & 6.124 & 0.205 & 0.028 & 0.126 & 1.30 & 3.74 & 0.119 & 4.616 & 0.284 & 0.028 & 0.129 & 1.35 & 3.68 & 0.119 \\ 
        & 0.4 & 4.673 & 0.233 & 0.022 & 0.121 & 1.26 & 3.52 & 0.112 & 4.441 & 0.187 & 0.025 & 0.124 & 1.31 & 3.58 & 0.116 \\ 
        & 1.6 & 4.219 & 0.192 & 0.022 & 0.116 & 1.18 & 3.37 & 0.104 & 4.355 & 0.217 & 0.023 & 0.120 & 1.25 & 3.47 & 0.110 \\ 
        & 6.4 & 4.473 & 0.207 & 0.019 & 0.112 & 1.12 & 3.24 & 0.102 & 4.291 & 0.208 & 0.023 & 0.121 & 1.26 & 3.49 & 0.111 \\ 
        & 12.8 & 4.407 & 0.117 & 0.018 & 0.113 & 1.12 & 3.25 & 0.102 & 4.639 & 0.221 & 0.024 & 0.123 & 1.25 & 3.54 & 0.113 \\ 
      \bottomrule
      \end{tabular}%
    }
    \caption{Target City Fine-tune Experiment Results}
    \label{tab:finetune_total}%
\end{table*}%

\begin{table*}[t]
    \centering
    \resizebox{1.0\textwidth}{!}{
      \begin{tabular}{c|c|ccccccc|ccccccc}
      \toprule
        \multirow{2}{*}{Target} & \multirow{2}{*}{Method} & \multicolumn{7}{c|}{Source City XA for BJ, BJ for CD and CD for XA} & \multicolumn{7}{c}{Source City BJ for XA, XA for CD and CD for BJ} \\
        \cmidrule{3-16}
         &  & \makecell{Distance \\ ($\times 10^{-3}$)} & \makecell{Radius \\ ($\times 10^{-3}$) } & LocFreq & Hausdorff & DTW  & EDT & EDR & \makecell{Distance \\ ($\times 10^{-3}$)} & \makecell{Radius \\ ($\times 10^{-3}$)} & LocFreq & Hausdorff & DTW  & EDT & EDR \\
      \midrule

      \multirow{4}{*}{BJ} & w/o Cost & 1.505 & 1.190 & 0.082 & 0.373 & 6.99 & 12.47 & 0.248 & 1.704 & 2.969 & 0.120 & 0.444 & 9.99 & 15.62 & 0.290 \\ 
        & w/o Pref & 0.444 & 0.564 & 0.039 & 0.303 & 5.07 & 8.97 & 0.199 & 0.448 & 0.627 & 0.040 & 0.308 & 5.08 & 9.01 & 0.201 \\ 
        & w/o SS & 0.439 & 0.363 & 0.042 & 0.297 & 4.95 & 9.08 & 0.194 & 0.511 & 0.346 & 0.056 & 0.370 & 6.30 & 10.12 & 0.238 \\ 
        & \name & 0.636 & 0.111 & 0.041 & 0.292 & 4.89 & 8.73 & 0.190 & 0.643 & 0.182 & 0.043 & 0.315 & 5.27 & 9.17 & 0.205 \\ 
      \midrule
      \multirow{4}{*}{XA} & w/o Cost & 17.745 & 3.763 & 0.118 & 0.274 & 3.52 & 6.98 & 0.227 & 19.171 & 5.066 & 0.119 & 0.264 & 3.78 & 8.51 & 0.260 \\ 
        & w/o Pref & 4.641 & 1.052 & 0.062 & 0.219 & 3.00 & 5.03 & 0.164 & 4.356 & 0.293 & 0.047 & 0.198 & 2.54 & 4.61 & 0.146 \\ 
        & w/o SS & 5.261 & 2.817 & 0.059 & 0.228 & 3.15 & 4.95 & 0.163 & 3.685 & 0.673 & 0.041 & 0.199 & 2.57 & 4.49 & 0.144 \\ 
        & \name & 4.375 & 0.206 & 0.042 & 0.188 & 2.32 & 4.29 & 0.142 & 4.039 & 0.228 & 0.040 & 0.187 & 2.34 & 4.23 & 0.135 \\ 
      \midrule
      \multirow{4}{*}{CD} & w/o Cost & 6.403 & 2.483 & 0.074 & 0.198 & 2.47 & 6.77 & 0.195 & 4.957 & 0.217 & 0.037 & 0.140 & 1.61 & 3.94 & 0.127 \\ 
        & w/o Pref & 10.607 & 0.496 & 0.035 & 0.136 & 1.41 & 4.09 & 0.128 & 7.616 & 0.257 & 0.029 & 0.131 & 1.37 & 3.90 & 0.120 \\ 
        & w/o SS & 5.061 & 0.348 & 0.033 & 0.133 & 1.37 & 3.95 & 0.123 & 5.095 & 0.362 & 0.035 & 0.137 & 1.42 & 4.09 & 0.129 \\ 
        & \name & 5.109 & 0.233 & 0.027 & 0.117 & 1.19 & 3.59 & 0.109 & 4.902 & 0.248 & 0.026 & 0.125 & 1.34 & 3.63 & 0.117 \\ 
      \bottomrule
      \end{tabular}%
      }
    \caption{Ablation Study Results}
    \label{tab:ablation_total}%
\end{table*}%

\subsection{Details of Baselines}
To evaluate our method, the following trajectory generation models are selected as baselines.
\begin{itemize}
    \item Random Walk~\cite{node2vec}(RW): This method, as utilized in Node2Vec~\cite{node2vec}, involves simulating city mobility by performing random walks on the graph. We analyzed the trajectory length and the distribution characteristics of starting nodes in the dataset and then executed random walks on the graph to simulate city mobility.

    \item EPR Models: Including Density-EPR~\cite{epr_1}(DE) and Spatial-EPR~\cite{epr_2}(SE). These models characterize human behavior into two patterns: Explore and Preferential Return. They introduce gravity models to simulate the impact of group mobility on individuals. Empirical parameters are set to accomplish sampling.
    
    \item TrajGen~\cite{trajgen}(TG): This method utilizes a CNN-based GAN to generate the synthetic trajectory image. Then, it extracts locations from the image and uses a Seq2Seq model to infer the real trajectory sequence. 

    \item SeqGAN~\cite{SeqGAN}(SG): This method is the classical sequence generation method, which combines policy gradient with GAN to solve sequence generation problem.

    \item SVAE~\cite{SVAE}(SV): This method is the first to combine the variational autoencoder with the Seq2Seq model to generate mobility trajectory data.
    
    \item MoveSim~\cite{MoveSim}(MS): This method builds a self-attention-based trajectory generator and designs a mobility regularity-aware discriminator to train the generator in the reinforcement learning paradiagm.
    
    \item TS-TrajGen~\cite{ts_trajgen}(TT): This method combines the A* search algorithm with neural networks to model agent policy, and proposes a two-stage adversarial generative network to efficiently generate trajectory data.
    
    \item DiffTraj ~\cite{difftraj}(DT): This method utilize the generative abilities of diffusion model to reconstruct and synthesize geographic trajectories from white noise through a reverse trajectory denoising process.
    
    \item VOLUNTEER ~\cite{volunteer}(VO): This method employs Variational Autoencoder (VAE) to model the complex spatio-temporal distribution of users and obtain accurate trajectory simulations.
\end{itemize}
In the deeplearning methods, DiffTraj and VOLUNTEER do not incorporate travel demand information as input to the model. SeqGAN, SVAE, and MoveSim only take the starting road segment as input to the model. In contrast, TS-TrajGen and our model take both the starting and destination road segments from the travel demand as input, making them more suitable for simulating trajectory data for given travel demands. 

\subsection{Evaluation Metrics}
\label{apdx:metrics}
In the macro-similarity perspective, what we focus on is the overall statistical distribution of the trajectory dataset.
We evaluate the quality of the generated data by comparing the similarity of mobility patterns and urban traffic state metrics between the generated and real data. To obtain quantitative results, we use Jensen-Shannon Divergence(JS-Divergence) to calculate the similarity of real trajectory dataset $\mathcal{T}$ and generated trajectory dataset $\mathcal{\hat{T}}$, as follows
\begin{equation}
    Sim^{(mac)} = \text{JSD}(P(\mathcal{T}), P(\mathcal{\hat{T}})).
\end{equation}
In detail, we calculate the JS-Divergence in the following aspects.
\begin{itemize}
    \item Distance: Travel distance, which represents the spatial length of a trajectory.
    \item Radius: Radius of gyration, which represents the spatial travel range of a trajectory.
    \item LocFreq: Visit frequency distribution of each single road segment, which indicates the popularity of roads.
\end{itemize}

In the micro-similarity perspective, we focus on measuring the
sequence distance between the real trajectories and the generated trajectories with the same travel demand, as follows
\begin{equation}
    Sim^{(mic)} = \frac{1}{N} \sum_{\tau \in \mathcal{T}, \hat{\tau} \in \hat{\mathcal{T}}} \text{SeqDist}(\tau, \hat{\tau}). 
\end{equation}
In practice, we use four widely used trajectory distance metrics to
measure the micro similarity.
\begin{itemize}
    \item Hausdorff distance: The Hausdorff distance quantifies the maximum distance from any point in one set to the nearest point in another set.
    \item DTW: It finds an optimal alignment between sequences by warping the time axis to minimize the cumulative distance between corresponding points.
    \item EDT: It calculates the minimum edit distance between two sequence. 
    \item EDR: Edit distance on real sequence, a variant of editing distance.
\end{itemize}

\subsection{Model Settings}
All experiments were conducted on a machine equipped with an NVIDIA GeForce 3090 GPU, running the Ubuntu 20.04 operating system. The model was implemented using Pytorch 1.12.1. The number of SAGAT layer is $6$. During 
training, the number of clusters $K$ varies according to the size of the dataset; a batch of data consists of $k=3$ clusters; the learning rate is set to $10^{-5}$; and the training epochs is $600$. The balancing weights in the loss functions are: $\lambda_r = 50$, $\lambda_d = 100$ and $\lambda_g = 5$.

\section{Experiment Result}
\label{apdx:result}
We completed six groups of experiments on three datasets, with each of the three cities serving as both the source and target city in different combinations, to evaluate the effectiveness of fine-tune in the target cities and the impact after ablating key submodules. The complete results of the target city fine-tune experiment are shown in the Table~\ref{tab:finetune_total} and the complete results of ablation study are shown in the Table~\ref{tab:ablation_total}.

\end{document}